\documentclass[conference]{IEEEtran}
\IEEEoverridecommandlockouts
\usepackage{cite}
\usepackage{amsmath,amssymb,amsfonts}
\usepackage{algorithmic}
\usepackage{graphicx}
\usepackage{textcomp}
\usepackage{xcolor}
\usepackage{booktabs,multirow,subfigure,algorithm,hyperref}

\def\BibTeX{{\rm B\kern-.05em{\sc i\kern-.025em b}\kern-.08em
    T\kern-.1667em\lower.7ex\hbox{E}\kern-.125emX}}
\begin{document}

\title{Customizing Synthetic Data for Data-Free Student Learning}

\author{\IEEEauthorblockN{Shiya Luo}
\IEEEauthorblockA{\textit{Zhejiang University} \\
Hangzhou, China \\
lsya@zju.edu.cn}
\and
\IEEEauthorblockN{Defang Chen}
\IEEEauthorblockA{\textit{Zhejiang University} \\
Hangzhou, China \\
defchern@zju.edu.cn}
\and
\IEEEauthorblockN{Can Wang}
\IEEEauthorblockA{\textit{Zhejiang University} \\
Hangzhou, China \\
wcan@zju.edu.cn}
}

\maketitle

\begin{abstract}
Data-free knowledge distillation (DFKD) aims to obtain a lightweight student model without original training data. Existing works generally synthesize data from the pre-trained teacher model to replace the original training data for student learning. To more effectively train the student model, the synthetic data shall be customized to the current student learning ability. However, this is ignored in the existing DFKD methods and thus negatively affects the student training. To address this issue, we propose Customizing Synthetic Data for Data-Free Student Learning (CSD) in this paper, which achieves adaptive data synthesis using a self-supervised augmented auxiliary task to estimate the student learning ability. Specifically, data synthesis is dynamically adjusted to enlarge the cross entropy between the labels and the predictions from the self-supervised augmented task, thus generating \textit{hard samples} for the student model. The experiments on various datasets and teacher-student models show the effectiveness of our proposed method. Code is available at: \href{https://github.com/luoshiya/CSD}{https://github.com/luoshiya/CSD}
\end{abstract}

\begin{IEEEkeywords}
data-free knowledge distillation, self-supervision, model compression
\end{IEEEkeywords}

\section{Introduction}
\label{sec:intro}

In recent years, convolutional neural networks (CNNs) have achieved remarkable success in various applications \cite{simonyan2015Very,He2016Deep,Zagoruyko2016Wide} with over-parameterized architectures. But its expensive storage and computational costs make model deployment on mobile devices difficult. Therefore, knowledge distillation (KD) \cite{hinton2015distilling,chen2022knowledge} comes into play to compress models by transferring dark knowledge from a well-trained cumbersome teacher model to a lightweight student model. The prevailing knowledge distillation methods \cite{hinton2015distilling,Romero2015FitNets,chen2020online,Chen2021Cross,wang2023semckd} depend on a strong premise that the original data utilized to train the teacher model is directly accessible for student training. However, this is not always the case in some practical scenarios where the data is not publicly shared due to privacy, intellectual property 
concerns or excessive data size etc. Data-free knowledge distillation (DFKD) \cite{Chen2019DataFreeLO} is thus proposed to solve this problem.

\begin{figure}[t]
	\centering
	\begin{subfigure}[Traditional adversarial framework\label{fig:compareA}]{
		\centering
		\includegraphics[width=0.92\linewidth]{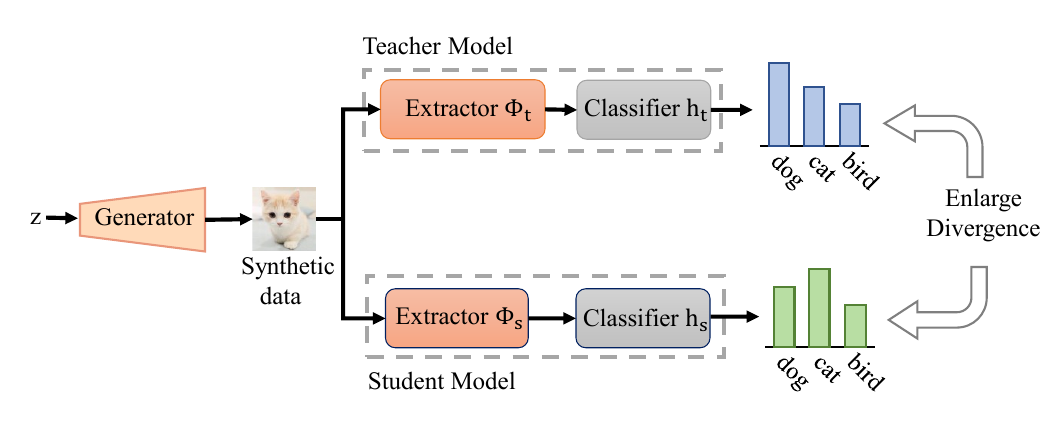}}
	\end{subfigure}
	\centering
	\begin{subfigure}[Our proposed  adversarial framework\label{fig:compareB}]{
		\centering
		\includegraphics[width=0.92\linewidth]{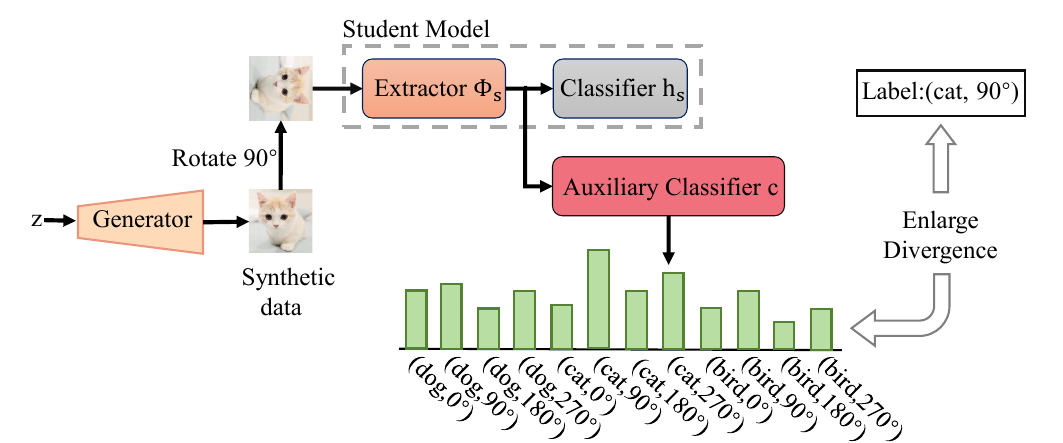}}
	\end{subfigure}
	\caption{Two different adversarial frameworks in hard data synthesis. (a) Traditional adversarial framework aims to enlarge divergence between predictions of the teacher and student. (b) Our proposed adversarial framework aims to enlarge divergence between labels and predictions from the self-supervised augmented task given to the student.}
	\label{fig:comparison}
\end{figure}

Existing DFKD methods generally divide each training round into two stages: \textit{data synthesis} and \textit{knowledge transfer}. Two different approaches are proposed in the data synthesis stage: \textit{model inversion} inputs the random Gaussian noise into the fixed teacher model and iteratively updates the input via the back-propagation from the teacher model \cite{Yin2020DreamingTD,fang2021contrastive}; \textit{generative reconstruction} utilizes a generator network to learn a mapping from the low-dimensional noise to the desired high-dimensional data manifold close to the original training data \cite{Chen2019DataFreeLO,Yoo2019KnowledgeEW, hao2021model}. In the knowledge transfer stage, the synthetic data from the previous stage is used to train the student model with the regular knowledge distillation procedure.

As training progresses, easy samples bring little new knowledge and contribute less to the student learning. The key to improvement of the student learning ability is to provide the student with hard samples in training such that it can continuously acquire new knowledge. 
Some existing adversarial DFKD methods generate hard samples on which the student disagree with the teacher by enlarging the divergence between their prediction distribution \cite{Micaelli2019ZeroshotKT,choi2020data,fang2021contrastive,fang2022up} (see Fig.~\ref{fig:compareA}). However, the teacher has not been trained on such synthetic samples, and thus soft predictions for many samples are likely to be inaccurate. The student will experience minimal improvement, or even a decline, in its learning ability when attempting to imitate the teacher on those incorrect samples (as shown in Fig.~\ref{fig:acc_curve}). Furthermore, it is difficult to manually evaluate whether soft predictions of the teacher is correct.

In this paper, we propose Customizing Synthetic Data for Data-Free Student Learning (CSD), which directly takes the current student learning ability as a reference to adaptively synthesize hard samples and the learning ability is estimated through a self-supervised augmented auxiliary task that learns the joint distribution of the classification task and the self-supervised rotation task. In this way, the capability of capturing semantic information can serve as a good indicator of the student learning ability, and the auxiliary task can effectively verify how well the student understand semantics \cite{gidaris2018unsupervised}. 
An extra auxiliary classifier appended to the student feature extractor learns the self-supervised augmented auxiliary task in knowledge transfer stage and then estimates the current student learning ability as an evaluator in data synthesis stage by calculating the divergence between labels and predictions from the auxiliary task. In this way, we accurately generate hard samples relative to current student learning ability by enlarging this divergence in an adversarial way. Different from the traditional adversarial objective \cite{Micaelli2019ZeroshotKT,choi2020data,fang2021contrastive,fang2022up}, we use the student model itself rather than the pre-trained teacher model to estimate the sample difficulty of the synthetic data (see Fig.~\ref{fig:compareB}), which is more reliable for the student training and beneficial for the student performance improvement. As shown in Fig.~\ref{fig:acc_curve}, the student improves its learning ability with our hard samples and are not easily disturbed by the teacher misinformation.

Our contributions are summarized as follows:

\begin{itemize}
\item We propose a novel method to dynamically generate hard samples based on the current learning ability of the student in the data-free knowledge distillation scenario.
\item An auxiliary classifier is used to learn a self-supervised augmented task, and also acts as an evaluator to estimate the student learning ability for hard data synthesis.
\item We conduct extensive experiments on various datasets and teacher-student model architectures. Experimental results confirm the effectiveness of our method.
\end{itemize}

\section{Proposed Method}
\begin{figure*}[t]
  \centering
  \includegraphics[width=0.95\linewidth]{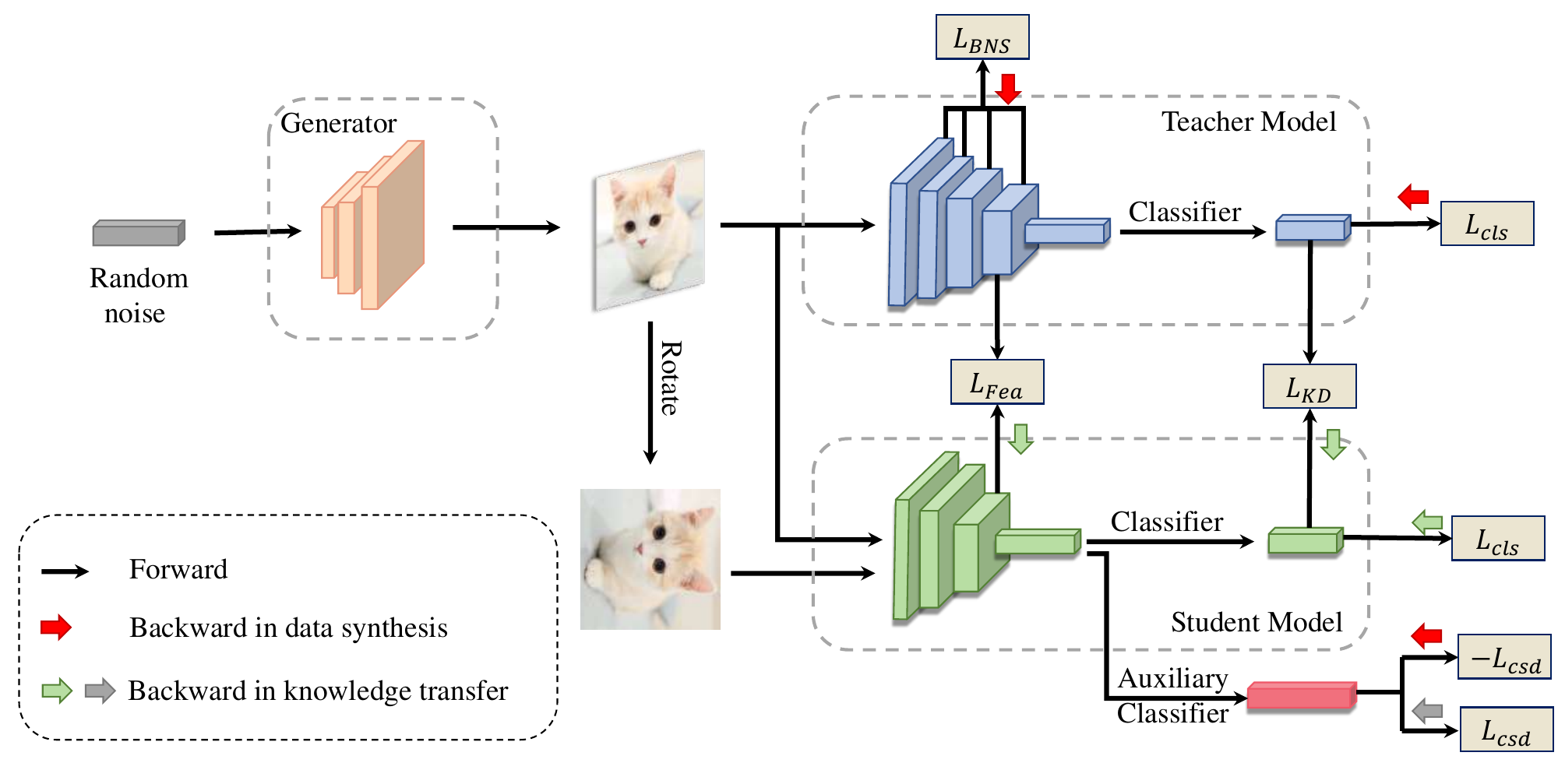}
  \caption{An overview of our proposed CSD. The student is equipped with an auxiliary classifier after feature extractor to predict the categories of rotated images with a self-supervised augmented task. 1) In data synthesis stage, random noise vector and the generator are jointly trained to synthesize images, not only based on outputs from the auxiliary classifier, but also the statistics of the teacher's BN layers and the given labels. 2) In knowledge transfer stage, the student is trained to imitate behaviors of the teacher, and the auxiliary classifier is trained separately to improve its own evaluating ability.}
  \label{fig:overview}
\end{figure*}

The overview of our proposed CSD framework is shown in Fig.~\ref{fig:overview}. The framework consists of a fixed pre-trained teacher, a generator, a student and an auxiliary classifier appended to the student feature extractor. 
The generator and the auxiliary classifier are trained in an adversarial manner. In data synthesis stage, the generator would explore hard samples based on the student learning ability with the auxiliary classifier. In knowledge transfer stage, the auxiliary classifier tries to improve its own evaluating ability. Two stages are executed alternately until convergence. 

\subsection{Data Synthesis}
In data synthesis stage, we follow CMI \cite{fang2021contrastive} to synthesize data $\tilde{x}\in\mathbb{R}^{H\times W \times C}$ (H, W, C denote the height, width and channel number, respectively) from a pre-trained teacher model as the surrogate for original training data $x$. We jointly update random noise vector $z$ and the parameters $\theta_{g}$ of the generator $\mathcal{G}$ to obtain $\tilde{x}=\mathcal{G}\left(z\right)$ for $n_g$ steps in each training round. The generator provides stronger regularization on pixels due to the shared parameters $\theta_{g}$. 

Although the main purpose of our work is to synthesize hard data based on the current ability of the student itself, if we synthesize data only by the student, this may make the distribution of the synthetic data far away from the original training data due to the lack of data prior constraints. The optimization objective of data synthesis consists of two components and is formulated as:
\begin{equation}
  \min_{z,\theta_{g}} \mathcal L_{narrow}-\alpha\mathcal L_{csd},
  \label{eq:ds}
\end{equation}
where $\mathcal{L}_{narrow}$ aims to narrow the gap between the synthetic data and the original training data with the help of the well-trained teacher model for alleviating outliers, and $\mathcal{L}_{csd}$ estimates the learning ability of the student. We will elaborate these two terms later.

{\bf Narrowing the Distribution Gap.} To make synthetic data more realistic, we adopt the following optimization objective to narrow the gap between the distribution of synthetic data and original training data:
\begin{equation}
    \mathcal{L}_{narrow} = \mathcal{L}_{cls} + \mathcal{L}_{bns},
    \label{eq:narrow}
\end{equation}
$\mathcal{L}_{cls}$ represents an one-hot assumption that if the synthetic data have the same distribution as that of the original training data, the prediction of the synthetic data by the teacher model would be like a one-hot vector \cite{Chen2019DataFreeLO}. Therefore, $\mathcal{L}_{cls}$ is calculated as the cross entropy between the teacher prediction $\mathcal{T}\left(\tilde{x}\right)$ and the pre-defined label $\tilde{y}$:
\begin{equation}
    \label{eq:cls}
    \mathcal{L}_{cls}=CrossEntropy\left(\tilde{y}, \mathcal{T}\left(\tilde{x}\right)\right),
\end{equation}
$\mathcal{L}_{bns}$ is a constraint that effectively utilizes statistics stored in the batch normalization (BN) layers of the teacher as data prior information \cite{Yin2020DreamingTD}. It employs running mean $\mu_{l}$ and running variance $\sigma_{l}^{2}$ of the $l$-th BN layer as feature statistics of original training data. $\mathcal{L}_{bns}$ is then calculated as the l2-norm distance between features statistics of synthetic data $\tilde{x}$ and original training data:
\begin{equation}
    \label{eq:bns}
    \mathcal{L}_{bns}=\sum_{l}\left(\Vert\tilde{\mu}_{l}(\tilde{x})-\mu_{l}\Vert_2+\Vert\tilde{\sigma}_l^2(\tilde{x})-\sigma_l^2\Vert_2\right),
\end{equation}
where $\tilde{\mu}_{l}(\tilde{x})$ and $\tilde{\sigma}_l^2(\tilde{x})$ are mean and variance of the feature maps at the $l$-th teacher layer, respectively.

{\bf Customizing Synthetic Data for the Student.} 
In each training round, it is necessary to synthesize data adaptively according to the current student learning ability, so as to prevent the student from repeatedly learning oversimple samples.  
To quantify learning ability, we consider that if a model can understand the semantic information of a image well, it would have a strong learning ability. Specifically, we adopt a simple self-supervised task by first rotating each image at different angles and then forcing the model to identify which angle each image comes from. As illustrated in \cite{gidaris2018unsupervised}, the model can effectively perform the rotation recognition task unless it first learns to recognize the object categories and then recognize semantic parts in the image. 
But only using the rotation task to estimate learning ability is not enough. For example,``6'' is rotated $180^\circ$ for the digit ``9'' and $0^\circ$ for the digit ``6''. Inspired by \cite{Yang2021HierarchicalSA}, we also combine the original classification task and the self-supervised rotation task into a unified task, named as the self-supervised augmented task, which forces the model to identify the angle as well as the category to eliminating incorrect estimation.

We consider a N-way classification task and a M-way self-supervised rotation task. The CNN student model consists of two components: the feature extractor $\Phi:\tilde{x}\rightarrow\mathbb{R}^d$ and the classifier $h:\mathbb{R}^d\rightarrow\mathbb{R}^N $, i.e., $\mathcal{S}(\tilde{x})=h(\Phi(\tilde{x}))$. Here d denotes the feature dimension. we attach an auxiliary classifier $c:\mathbb{R}^d\rightarrow\mathbb{R}^K $ with parameters $\theta_c$ behind the feature extractor, where $K=N*M$ represents the number of categories for the self-supervised augmented task. $\mathcal{L}_{csd}$ is calculated as follows:
\begin{equation}
    \label{eq:csd}
    \mathcal{L}_{csd} = CrossEntropy\left(k, c\left(\Phi\left(trans\left(\tilde{x}\right)\right)\right)\right),
\end{equation}
where $trans(\cdot)$ is the operation of rotation and k is the label of the rotated version of synthetic data $\tilde{x}$ in the self-supervised augmented task. For example, if the category of $\tilde{x}$ in the original classification task is $n$ and the category of its rotated version in the self-supervised rotation task is $m$, then the category in the self-supervised augmented task is $n*M+m$. By enlarging $\mathcal{L}_{csd}$, we generate hard samples on which the student has difficulty understanding semantics. 

\begin{algorithm}[t] 
\caption{Self-Supervised Data-Free Knowledge Distillation} 
\label{alg:Framwork} 
\textbf{Input}: A pretrained teacher model $\mathcal{T}\left(\tilde{x};\theta_{t}\right)$; A randomly initialized student model $\mathcal{S}\left(\tilde{x};\theta_{s}\right)$;  An initialized image bank $\mathcal{B}$. \\
\textbf{Output}: A tiny student model $\mathcal{S}\left(\tilde{x};\theta_{s}\right)$. 

\begin{algorithmic}[1] 
\FOR{$i=1$ to $epochs$}
\STATE
//stage 1: data synthesis\\
\STATE
initialize the generator  $\mathcal{G}\left(z;\theta_{g}\right)$
\STATE
$z\leftarrow\mathcal{N}\left(0, I\right)$
\FOR {$j=1$ to $n_g$ }
\STATE
$\tilde{x}\leftarrow \mathcal{G}\left(z;\theta_g\right)$
\STATE
$\mathcal L_{DS}=\mathcal L_{narrow}-\alpha\mathcal L_{csd}$
\STATE
$z\leftarrow z-\eta\nabla_z\mathcal L_{DS}$
\STATE
$\theta_g\leftarrow\theta_g-\eta\nabla_g\mathcal L_{DS}$
\ENDFOR
\STATE
$\mathcal{B}\leftarrow \mathcal{B}\cup \tilde{x}$
\STATE
//stage 2: knowledge transfer \\
\STATE
initialize the auxiliary classifier $\mathcal{C} \left( \phi;\theta_{c}\right)$
\FOR{$j=1$ to $n_s$}
\STATE
sample $\tilde{x}$ from $\mathcal{B}$
\STATE
$\mathcal L_{KT}= \mathcal{L}_{ce}+\mathcal{L}_{kd}+\beta*\mathcal{L}_{fea}$
\STATE
$\theta_s\leftarrow\theta_s-\xi\nabla_s\mathcal L_{KT}$
\STATE
$\theta_c\leftarrow \theta_c-\xi\nabla_c\mathcal L_{csd}$
\ENDFOR

\ENDFOR
\end{algorithmic}
\end{algorithm}

\begin{table*}[t]
\caption{Top-1 test accuracy comparison on three datasets: SVHN, CIFAR-10 and CIFAR-100}
  \resizebox{0.98\linewidth}{!}{
  \begin{tabular}{cccccccccc}
    \toprule
    \multirow{2}{*}{Dataset} & \multirow{2}{*}{Teacher} & \multirow{2}{*}{Student} & \multicolumn{7}{c}{Accuracy}\\
    \cmidrule{4-10}
    & & &  Teacher & Student & DAFL & ZSKT & ADI & CMI & CSD \\
    \midrule
    \multirow{5}{*}{SVHN} & WRN-40-2 & WRN-16-1 & 96.14\% & 95.27\% & 92.88\% &93.39\% & 87.15\% & 94.02\% & {\bf 94.57\%} \\
    & WRN-40-2 & WRN-40-1 & 96.14\% &  95.88\% & 94.78\% &94.84\% & 89.16\% & 94.82\% & {\bf 95.58\%} \\
    & WRN-40-2 & VGG8 & 96.14\% & 94.60\% & 74.78\% &89.16\% & 86.94\% & 90.29\% & {\bf 92.72\%} \\
    & WRN-40-2 & MobileNet-V2 & 96.14\% & 95.24\% & 79.34\% & 91.56\% & 79.11\% & 91.49\% & {\bf 91.89\%} \\
    & ResNet34 & ResNet18 & 95.62\% & 95.17\% & 94.57\% & 94.48\% & 82.57\% & 94.82\% & {\bf 95.06\%} \\
    \midrule
    \multirow{5}{*}{CIFAR-10} & WRN-40-2 & WRN-16-1 & 94.87\% & 91.12\% & 68.97\% &80.91\% & 74.77\% & 88.81\% & {\bf 90.50\%} \\
    & WRN-40-2 & WRN-40-1 & 94.87\% & 93.94\% & 77.87\% & 85.41\% & 84.63\% & 92.37\% & {\bf 93.02\%} \\
    & WRN-40-2 & VGG8 & 94.87\% & 91.28\% & 53.3\% & 47.75\% & 59.04\% & 87.66\% & {\bf 88.57\%} \\
    & WRN-40-2 & MobileNet-V2 & 94.87\% & 89.29\% & 43.12\% & 23.39\% & 52.52\% & 82.50\% &  {\bf 82.95\%} \\
    & ResNet34 & ResNet18 & 95.70\% & 95.20\% & 89.07\% & 90.90\% & 89.88\% & 94.38\% & {\bf 94.73\%} \\
    \midrule
    \multirow{5}{*}{CIFAR-100} & WRN-40-2 & WRN-16-1 & 75.83\% & 65.31\%  & 22.06\% & 30.15\%  & 35.99\% & 56.46\% & {\bf 60.88\%} \\
    & WRN-40-2 & WRN-40-1 & 75.83\% & 72.19\% & 38.49\% &39.51\%  & 39.46\% & 68.62\% & {\bf 69.69\%} \\
    & WRN-40-2 & VGG8 & 75.83\% & 68.76\% & 25.24\% &10.08\%  & 32.17\% & 64.50\% & {\bf 66.15\%} \\
    & WRN-40-2 & MobileNet-V2 & 75.83\% & 62.38\% & 22.02\% & 4.42\%   & 18.81\% & 56.79\% & {\bf 59.95\%} \\
    & ResNet34 & ResNet18 & 78.05\% & 77.10\% & 67.91\% & 61.32\% & 57.75\% & 75.06\% & {\bf 76.03\%} \\
    \bottomrule
  \end{tabular}}
  
  \label{tab:comparison}
\end{table*}

\subsection{Knowledge Transfer}
In knowledge transfer stage, the main purpose is to encourage the student model to mimic behaviors of the teacher model. The vanilla KD \cite{hinton2015distilling} matches final prediction distribution of the teacher and student model by calculating the Kullback-Leibler (KL) divergence between outputs of the teacher and the student:
\begin{equation}
    \label{eq:kd}
    \mathcal{L}_{kd} = KL\left(\sigma\left(\mathcal{T}\left(\tilde{x}\right)/\tau\right), \sigma\left(\mathcal{S}\left(\tilde{x}\right)/\tau\right)\right),
\end{equation}
where $\sigma(\cdot)$ is the softmax function and $\tau$ is a hyper-parameter to soften the distribution. We set $\tau$ to 20 throughout all experiments for fair comparison as CMI \cite{fang2021contrastive}. 

Besides prediction distribution, feature maps can also be used as valuable knowledge to effectively guide the student \cite{Romero2015FitNets}. We define the Mean-Square-error (MSE) loss between teacher feature maps $F_t\in\mathbb{R}^{H_t*W_t*C_t}$ and student feature maps $F_s\in\mathbb{R}^{H_s*W_s*C_s}$ from the last layer as:
\begin{equation}
    \label{eq:fea}
    \mathcal{L}_{fea} = MSE(F_t, r(F_s)),
\end{equation}
where $r(\cdot)$ is a projection to align the dimension of feature maps. The student is trained for $n_s$ steps in each training round and optimized by:
\begin{equation}
    \label{eq:kt}
    \min_{\theta_{s}}  \mathcal{L}_{ce}+\mathcal{L}_{kd}+\beta*\mathcal{L}_{fea},
\end{equation}
where $\beta$ is a hyper parameter to balance the three loss items, and $\mathcal{L}_{ce}=CrossEntropy(\tilde{y},\mathcal{S}(\tilde{x}))$ is a regular loss in the original classification task to calculate cross entropy between student outputs and pre-defined labels.

Besides the student training, the auxiliary classifier is also separately trained with the following loss to improve its own evaluation capability to better help the data synthesis stage:
   \begin{equation}
   \label{eq:t_csd}
    \min_{\theta_{c}} \mathcal{L}_{csd}.
\end{equation} 

\subsection{Training Procedure}
The two-stage training procedure is summarized in Algorithm~\ref{alg:Framwork}. In the data synthesis stage, the random noise $z$ and generator $\mathcal{G}$ are first trained for $n_g$ times. Then we append the new synthetic data into an image bank for preventing catastrophic forgetting \cite{Binici2022PreventingCF,Binici2022RobustAR}. In knowledge transfer stage, we sample data from the image bank and separately train the student $\mathcal{S}$ and the auxiliary classifier $c$ for $n_s$ times. 

\section{EXPERIMENTS}
{\bf Datasets and models.} We conduct experiments on SVHN \cite{Netzer2011ReadingDI}, CIFAR-10 and CIFAR-100 \cite{krizhevsky2009learning} datasets, following a similar training setting as \cite{fang2021contrastive}. For all datasets, various models are used, including ResNet \cite{He2016Deep}, WRN \cite{Zagoruyko2016Wide}, VGG \cite{simonyan2015Very} and MobileNet \cite{Sandler2018MobileNetV2}. The generator architecture is the same as \cite{Chen2019DataFreeLO}. 

{\bf Training details.} For all datasets, to prevent the student from overfitting to data generated by early training rounds \cite{Binici2022PreventingCF,Binici2022RobustAR}, we first synthesize some data to initialize the image bank by removing $\mathcal{L}_{csd}$ and running 400 synthesis batches with each one containing 200 samples. We totally train 100 rounds (epochs). In data synthesis stage, the random noise vector and generator are updated using Adam optimizer with 1e-3 learning rate. We synthesize 200 images in each step and repeat for $n_g=500$ steps. The hyper-parameter $\alpha$ is set to 10. In knowledge transfer stage, the student and the auxiliary classifier are update using SGD optimizer with 0.1 learning rate, 0.9 momentum and 1e-4 weight decay and we adopt cosine annealing for the learning rate decay. we sample 128 images from the image bank in each step and repeat for $n_s=2000$ steps. The hyper-parameter $\beta$ is set to 30. We set temperature $\tau$ to 20. Test accuracy is used to evaluate the proposed method. We run all experiments for three times and report the means. More implementation details and results can be found in the appendix.

\subsection{Comparison with DFKD methods}
We compare with four representative DFKD methods on five groups of teacher-student models, including three homogeneous and two heterogeneous architecture combinations. DAFL \cite{Chen2019DataFreeLO} and ZSKT \cite{Micaelli2019ZeroshotKT} are generator-based methods. ADI \cite{Yin2020DreamingTD} and CMI \cite{fang2021contrastive} are inversion-based methods.

Table~\ref{tab:comparison} shows that our proposed CSD outperforms all other methods. We also observe that, except for CMI, other comparison methods perform poorly on heterogeneous combinations and more complex datasets. For example, in the case of ``WRN-40-2 \& VGG8" on CIFAR-100,  the test accuracy of DFAL is only 25.24\%, which do not even achieve half accuracy of the student trained on the original data (68.76\%). In contrast, our proposed CSD is robust on different datasets and teacher-student combinations.

\begin{table}[t]
    \centering
    \caption{Ablation study on WRN-40-2 \& WRN-16-1 to explore the effect of our proposed adversarial loss. Baseline denotes removing $\mathcal{L}_{csd}$. Adv denotes replacing $\mathcal{L}_{csd}$ with traditional adversarial loss $\mathcal{L}_{adv}$. Rotation denotes only adopting self-supervised rotation task.}
    \begin{tabular}{c|cc}
       \toprule
        Method & CIFAR-10 & CIFAR-100  \\
       \midrule
        Baseline & 86.88\% & 57.59\% \\
        Adv & 87.57\% & 53.5\% \\
        Rotation & 89.55\% & 59.32\% \\
        \midrule
       CSD & \bf90.50\% & \bf60.88\% \\
        \bottomrule
    \end{tabular}
    
    \label{tab:ss_module}
\end{table}


\begin{figure}[t]
	\centering	
     \begin{subfigure}[CIFAR-10\label{fig:acc_cifar10}]{
        \includegraphics[width=0.47\linewidth]{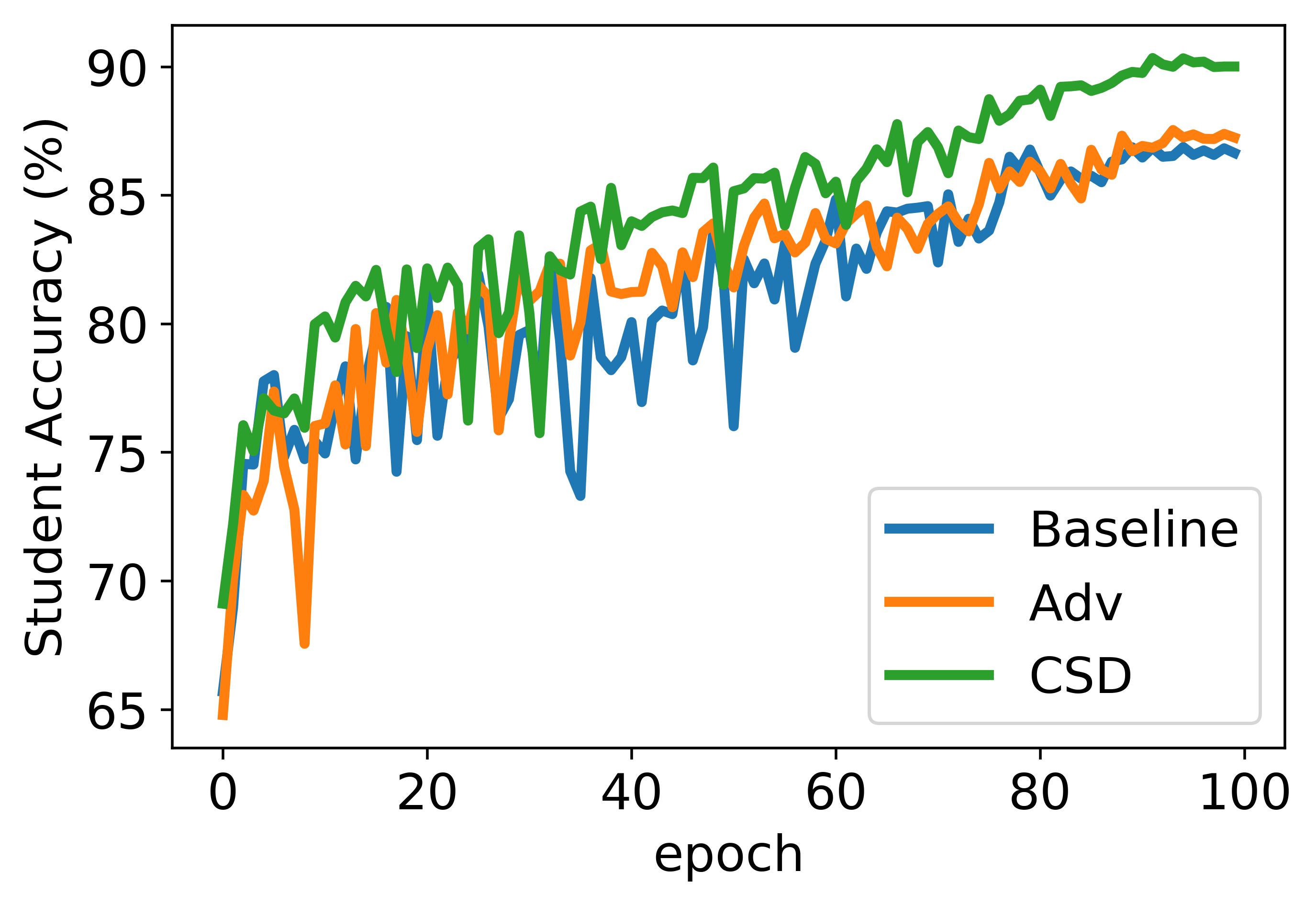}}
        \end{subfigure}
      \begin{subfigure}[CIFAR-100\label{fig:acc_cifar100}]{
        \includegraphics[width=0.47\linewidth]{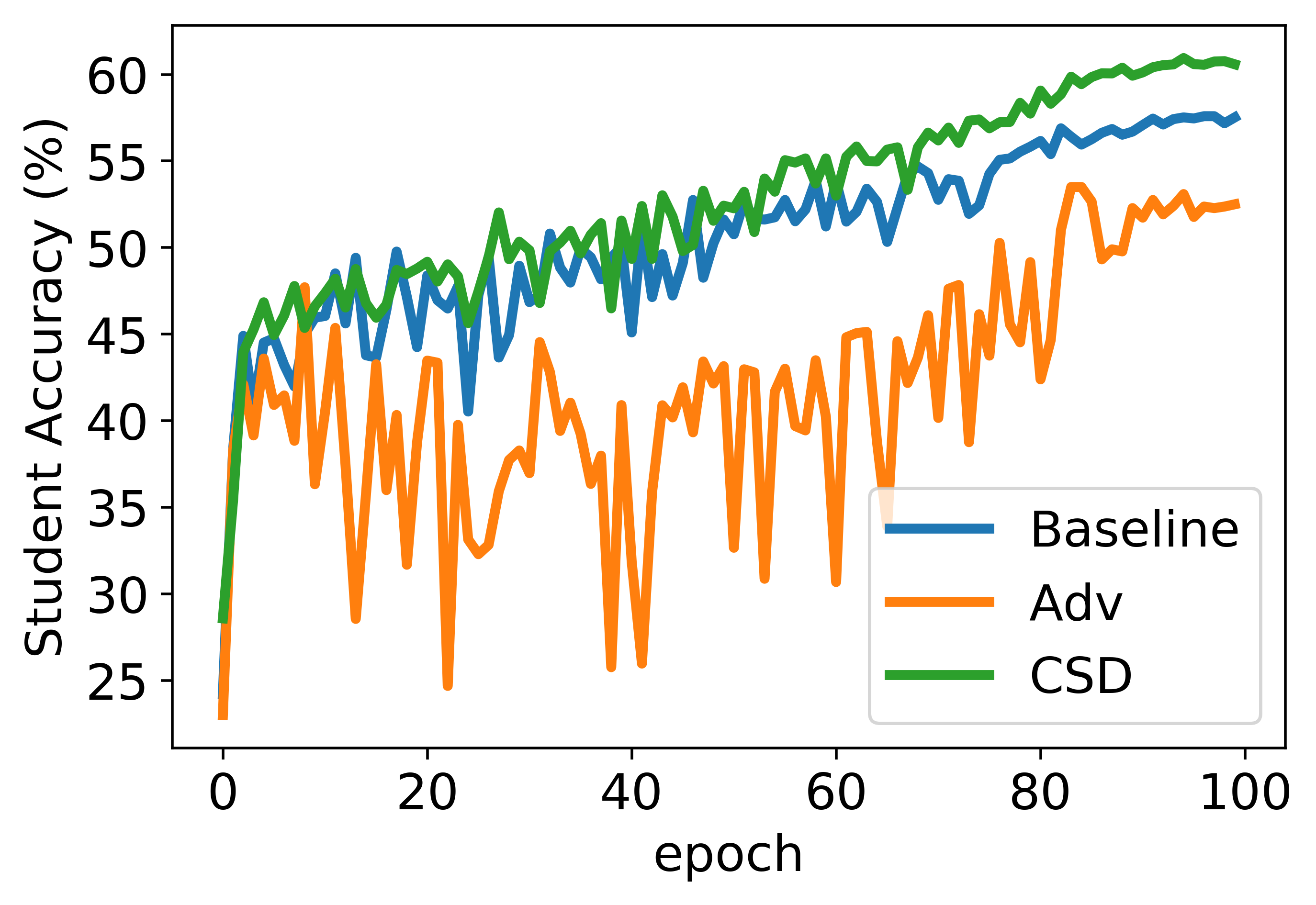}}
        \end{subfigure}
	\caption{Accuracy curves of the student trained by our CSD in comparison with baseline (removing $\mathcal{L}_{csd}$) and modified method (replacing $\mathcal{L}_{csd}$ with traditional adversarial loss $\mathcal{L}_{adv}$) on WRN-40-2 \& WRN-16-1.}
	\label{fig:acc_curve}
\end{figure}

\begin{figure}[t]
	\centering
    \centering
     \begin{subfigure}[CIFAR-10\label{fig:aux_cifar10}]{
        \includegraphics[width=0.47\linewidth]{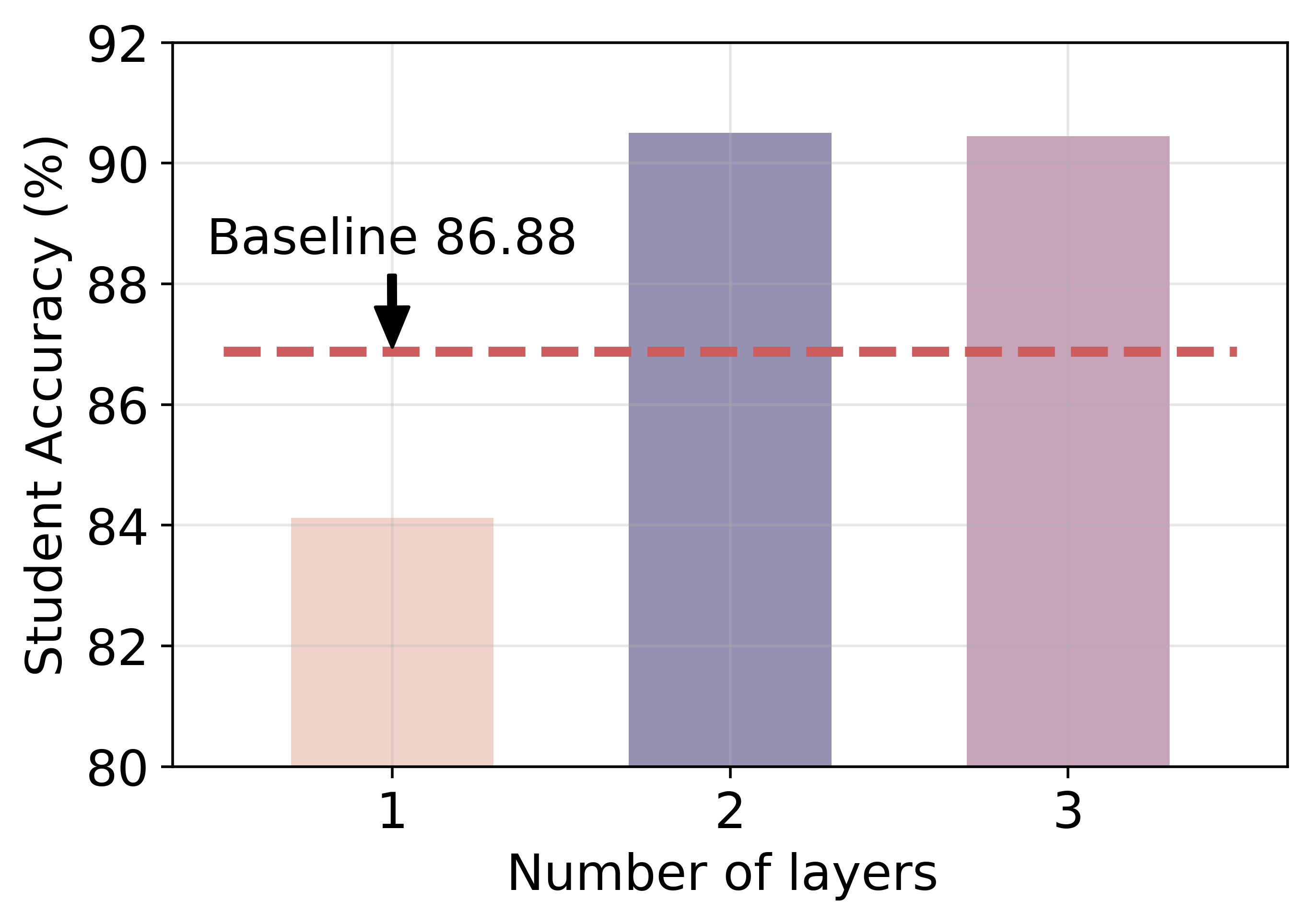}}
        \end{subfigure}
       \begin{subfigure}[CIFAR-100\label{fig:aux_cifar100}]{
        \includegraphics[width=0.47\linewidth]{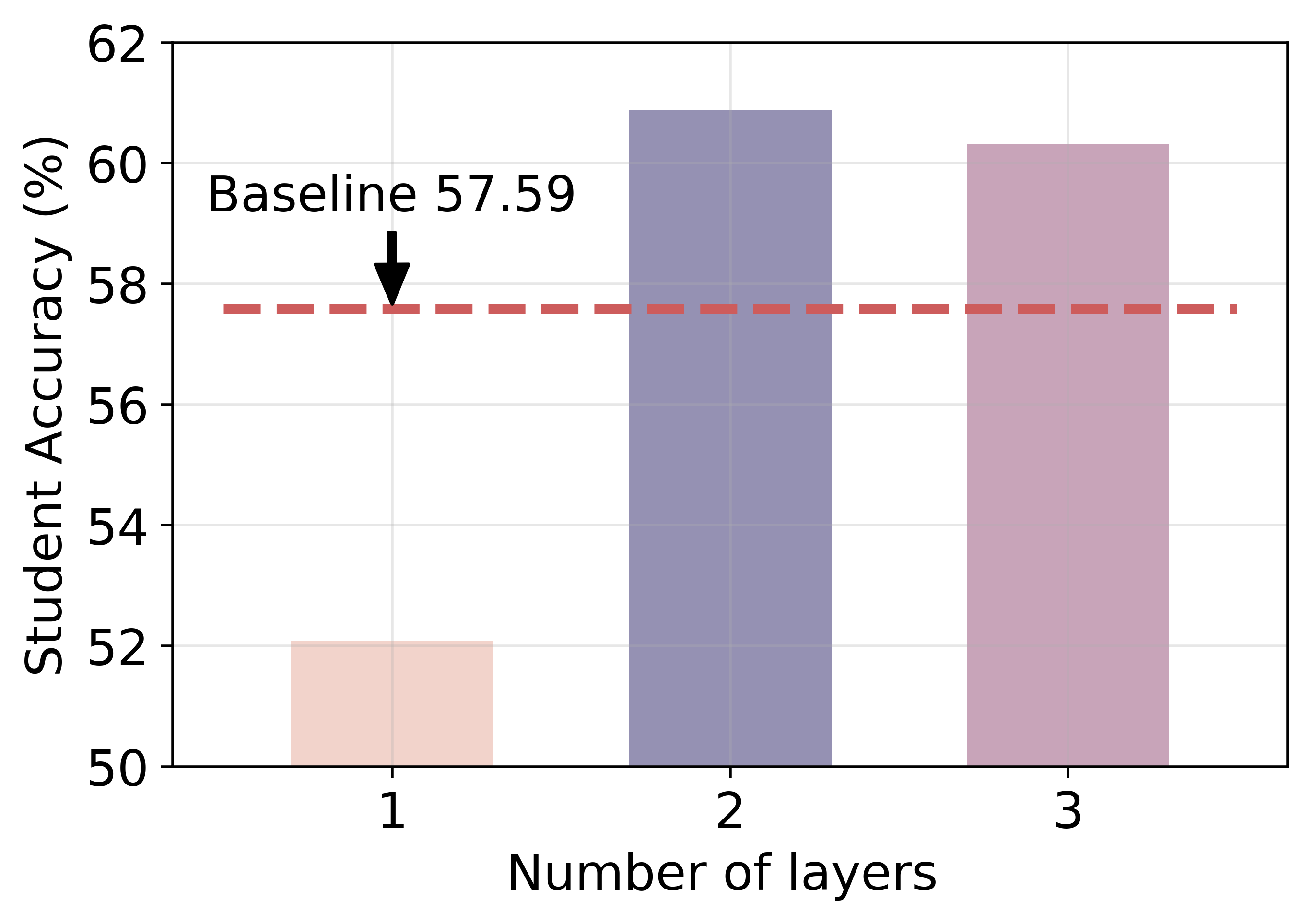}}
        \end{subfigure}
	\caption{Effect of the auxiliary classifier structure on WRN-40-2 \& WRN-16-1. The student is equipped with different numbers of fully-connected layers after feature extractor. }
	\label{fig:aux}
\end{figure}

\begin{table}[t]
    \centering
   \caption{Effect of training strategy of the student and auxiliary classifier during knowledge transfer on WRN-40-2 \& WRN-16-1.}
    \begin{tabular}{ccc}
       \toprule
        Method & CIFAR-10 & CIFAR-100  \\
       \midrule
        Baseline & 86.88\% & 57.59\% \\
        Joint training & 89.77\% & 60.43\% \\
        Separate training(CSD) & \bf90.50\% & \bf60.88\% \\
       \bottomrule
    \end{tabular}
    
    \label{tab:training}
\end{table}

\subsection{Effect of Our Proposed Adversarial Loss}
We conduct ablation study on CIFAR-10 and CIAFR-100 to explore whether our proposed adversarial loss $L_{csd}$ can help improve the student performance. As shown in Table~\ref{tab:ss_module}, in the case of Baseline, i.e., removing the adversarial loss (Equation \ref{eq:csd}), the accuracy drops by 3.62\% on CIFAR-10 (from 90.50\% to 86.88\%) and 3.29\% on CIFAR-100 (from 60.88\% to 57.59\%), which demonstrates the effectiveness of our proposed $\mathcal{L}_{csd}$.

To further demonstrate the superiority of our method, we compare with two alternative adversarial strategies. The first one is traditional adversarial manner as the previous work \cite{Micaelli2019ZeroshotKT,choi2020data,fang2021contrastive,fang2022up}, whose adversarial loss is to calculate the divergence between predictions of the teacher and student. We replace $\mathcal{L}_{csd}$ with traditional adversarial loss $ L_{adv} = KL\left(\sigma\left(\mathcal{T}\left(\tilde{x}\right)/\tau\right), \sigma\left(\mathcal{S}\left(\tilde{x}\right)/\tau\right)\right)$ and find that it has a slight improvement of 0.65\% (from 86.88\% to 87.57\%) compared to Baseline on CIFAR-10. Surprisingly, We observe that it even results in a large drop of 4.09\% (from 57.59\% to 53.5\%) on the more complex CIFAR-100 dataset. This indicates that estimating the sample difficulty with teacher predictions is likely to be unreliable, which would enlarge the negative effect in the case of teacher misdirection and thus weakens the student performance. Additionally, we plot the learning curves of the student trained by different strategies.
In Fig.~\ref{fig:acc_curve}, it is clear that $\mathcal{L}_{adv}$ causes very large accuracy fluctuations across training rounds (epochs), while our CSD makes the model converge faster and more stable.

The second alternative strategy is to use only the rotation task as the final task to quantify the student learning ability without containing the original classification task. So we replace $\mathcal{L}_{csd}$ with self-supervised rotation loss $\mathcal{L}_{rotation} = CrossEntropy\left(m,c\left(\Phi\left(trans\left(\tilde{x}\right)\right)\right)\right)$, where $m$ is the label of synthetic data in the rotation task. From Table~\ref{tab:ss_module}, this causes significantly performance improvement on both CIFAR-10 and CIFAR-100 compared to the traditional adversarial manner, which shows the superiority of synthesizing hard samples according to the current student learning ability. However, only rotation task may destroy the original visual semantic information on some samples (such as ``6'' vs ``9'') and results in inaccurate ability estimation. By combining the original classification task and the self-supervised rotation task, our CSD further improves the model performance. 


\subsection{Auxiliary Classifier Analysis}
Next, we explore how the structure and training strategy of the auxiliary classifier affect the final student performance. 

To study the effect of the auxiliary classifier structure, we attach different numbers of fully-connected layers (from 1 to 3) behind the feature extractor. In Fig.~\ref{fig:aux}, only one fully-connected layer even has a negative impact, which reduces the student performance on CIFAR-10 and CIFAR-100 by about 3\% and 5\% compared to the Baseline (without $\mathcal{L}_{csd}$), while two or three fully-connected layers can achieve similarly superior performance. We conjecture that multiple layers can effectively filter out noise in feature representations to accurately estimate the student ability. Therefore, we adopt two fully-connected layers as the auxiliary classifier for all experiments to trade off between the effectiveness and complexity. 

To study the effect of the training strategy during the knowledge transfer stage, we conduct experiments with two different training strategies: joint training and separate training.


(1) \textit{Joint training} updates the parameters of the student and the auxiliary classifier simultaneously at each step, that is, change the lines 17 and 18 of the Algorithm~\ref{alg:Framwork} to $\theta_s\leftarrow\theta_s-\xi\nabla_s(\mathcal L_{KT}+\mathcal L_{csd})$ and $\theta_c\leftarrow \theta_c-\xi\nabla_c\mathcal (\mathcal L_{KT}+\mathcal L_{csd})$. This strategy requires the student to learn the self-supervised augmented task together with the original classification task. 

(2) \textit{Separate training} is exactly our adopted strategy for CSD. At each step, we update the student parameters first and then fix it and turn to train the auxiliary classifier. 

Table~\ref{tab:training} demonstrates separate training performs better. We conjecture that the additional self-supervised auxiliary task might distract the student from the main classification task. 




\section{Conclusion}
In data-free knowledge distillation, the student model itself can act as a key contributor to synthesize more valuable data while this point is largely overlook previously. In this paper, we utilize a self-supervised augmented task to accurately estimate the current student learning ability in each training round to synthesize more valuable data rather than oversimple synthetic data. Extensive experiments are conducted on three popular datasets and various groups of teacher-student models to evaluate the performance of our proposed method, and the results demonstrates the effectiveness of our proposed CSD. A potential future work is to explore how to apply the popular diffusion models to synthetic samples for data-free knowledge distillation \cite{chen2023geometric}.


\section{Appendix}

\subsection{Experimental Details}

\subsubsection{Datasets}
We evaluate our proposed CSD on three public datasets for classification task: SVHN, CIFAR-10 and CIFAR-100. The details of these datasets are listed as follows:
\begin{itemize}
\item {\bf SVHN \cite{Netzer2011ReadingDI}.} SVHN is a dataset of street view house numbers collected by Google, and the size of each image is 32$\times$32. It consists of over 600,000 labeled images, including 73257 training images, 26,032 testing images and  531,131 additional training images. 
\item {\bf CIFAR-10 \cite{krizhevsky2009learning}.} CIFAR-10 is a dataset of 32$\times$32 colored images. It consists of 60,000 labeled images from 10 categories. Each category contains 6,000 images, which are divided into 5,000 and 1,000 for training and testing, respectively.
\item {\bf CIFAR-100 \cite{krizhevsky2009learning}.} CIFAR-100 is similar but more challenging to CIFAR-10, which consists of 100 categories. Each categories contains 500 training images and 100 testing images.
\end{itemize}

Note that the training set is only utilized for teacher training and is unseen for data-free knowledge distillation. However, the testing set is still used for assessment.

\subsubsection{Model Architectures}
For all datasets, three network types are used in teacher-student models: ResNet \cite{He2016Deep} ,WRN \cite{Zagoruyko2016Wide}, VGG \cite{simonyan2015Very} and MobileNet-V2 \cite{Sandler2018MobileNetV2}. The number behind ``VGG" and ``ResNet" denotes the depth of the network. ``WRN-n-k" denotes a residual network with $n$ depths and widening factor k. We use the same generator architecture as the previous work \cite{Chen2019DataFreeLO}, which is detailed in Table~\ref{tab:gen}. We set the dimension of random noise vector to 256. 

\begin{table}[t]
    \centering
        \caption{The architecture details of the generator. FC-c denotes a fully-connected layers with c units. Conv3-k denotes a convolutional layer with k 3 $\times$ 3 filters
    	and stride 1 × 1. Upsampling denotes a 2 $\times$ 2 nearest neighbour interpolation operation.}
    \label{tab:gen}
    \begin{tabular}{c}
    \toprule
        FC-8WH, reshape-(W/8,H/8,512), Batchnorm \\
        Upsampling, Conv3-128, Batchnorm, LeakyReLU \\
        Upsampling, Conv3-64, Batchnorm, LeakyReLU \\
        Conv3-3, Tanh, Batchnorm \\
    \bottomrule
    \end{tabular}
\end{table}

\subsubsection{Baseline}
We compare with four representative data-free knowledge distillation methods: two generator-based methods (DSFL and ZSKT) and two inversion-based methods (ADI and CMI). The details of these compared methods are listed as follows:
\begin{itemize}
    \item {\bf DAFL \cite{Chen2019DataFreeLO}.} DFAL is a generator-based DFKD method that introduces one-hot loss, activation loss and information entropy loss from the teacher feedback as constraints to generate data close to the original training data.
    \item {\bf ZSKT \cite{Micaelli2019ZeroshotKT}.} ZSKT is another generator-based DFKD method that first introduces adversarial distillation. It generate hard samples on which the student poorly matches the teacher, i.e., maximizing the KL divergence between their predictions, and then use these hard samples to minimize the KL divergence in order to train the student.
    \item {\bf ADI \cite{Yin2020DreamingTD}.} ADI is an inversion-based DFKD method that first proposes to utilize statistics stored in batch normalization layers of the teacher as image prior information.
    \item {\bf CMI \cite{fang2021contrastive}.} CMI is another inversion-based DFKD method that mainly addresses model collapse issue. It introduces a contrastive learning objective to encourage each sample to distinguish itself from others for sample diversity.
\end{itemize}

\begin{figure*}[t]
    \centering
  \begin{minipage}{0.5\linewidth}
    \rightline{
      \begin{minipage}{0.67\linewidth}
          \centering
          {\bf \large Epoch=1}
      \end{minipage} }
      \begin{minipage}{1\linewidth}
        \begin{minipage}{0.3\linewidth}
            \centering
            {\bf \large airplane}
        \end{minipage}
        \begin{minipage}{0.7\linewidth}
            \centering
            \includegraphics{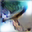}
            \includegraphics{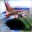}
            \includegraphics{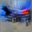}
            \includegraphics{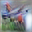}
            \includegraphics{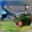}
        \end{minipage}
      \end{minipage}\\
      \begin{minipage}{1\linewidth}
        \begin{minipage}{0.3\linewidth}
            \centering
            {\bf \large horse}
        \end{minipage}
        \begin{minipage}{0.7\linewidth}
            \centering
            \includegraphics{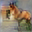}
            \includegraphics{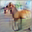}
            \includegraphics{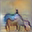}
            \includegraphics{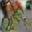}
            \includegraphics{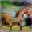}
        \end{minipage}
      \end{minipage}\\
      \begin{minipage}{1\linewidth}
        \begin{minipage}{0.3\linewidth}
            \centering
            {\bf \large ship}
        \end{minipage}
        \begin{minipage}{0.7\linewidth}
            \centering
          \includegraphics{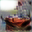}
            \includegraphics{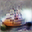}
            \includegraphics{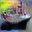}
            \includegraphics{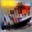}
            \includegraphics{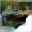}
        \end{minipage}\\
      \end{minipage}
  \end{minipage}
  \begin{minipage}{0.45\linewidth}
        \begin{minipage}{1\linewidth}
          \centering
          {\bf \large Epoch=50}
      \end{minipage} \\
      \begin{minipage}{1\linewidth}
            \centering
        \includegraphics{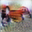}
        \includegraphics{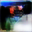}
        \includegraphics{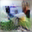}
        \includegraphics{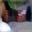}
        \includegraphics{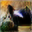}\\
        \includegraphics{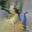}
        \includegraphics{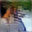}
        \includegraphics{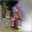}
        \includegraphics{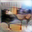}
        \includegraphics{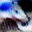}\\
        \includegraphics{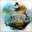}
        \includegraphics{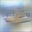}
        \includegraphics{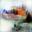}
        \includegraphics{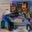}
        \includegraphics{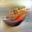}
      \end{minipage}
  \end{minipage}
    \caption{Visualization of images generated from different training epochs on CIFAR-10 for WRN-40-2 \& WRN-16-1}
    \label{fig:vis}
\end{figure*}


\begin{figure}[t]
	\centering	
     \begin{subfigure}[CIFAR-10\label{fig:ssacc_cifar10}]{
        \includegraphics[width=0.47\linewidth]{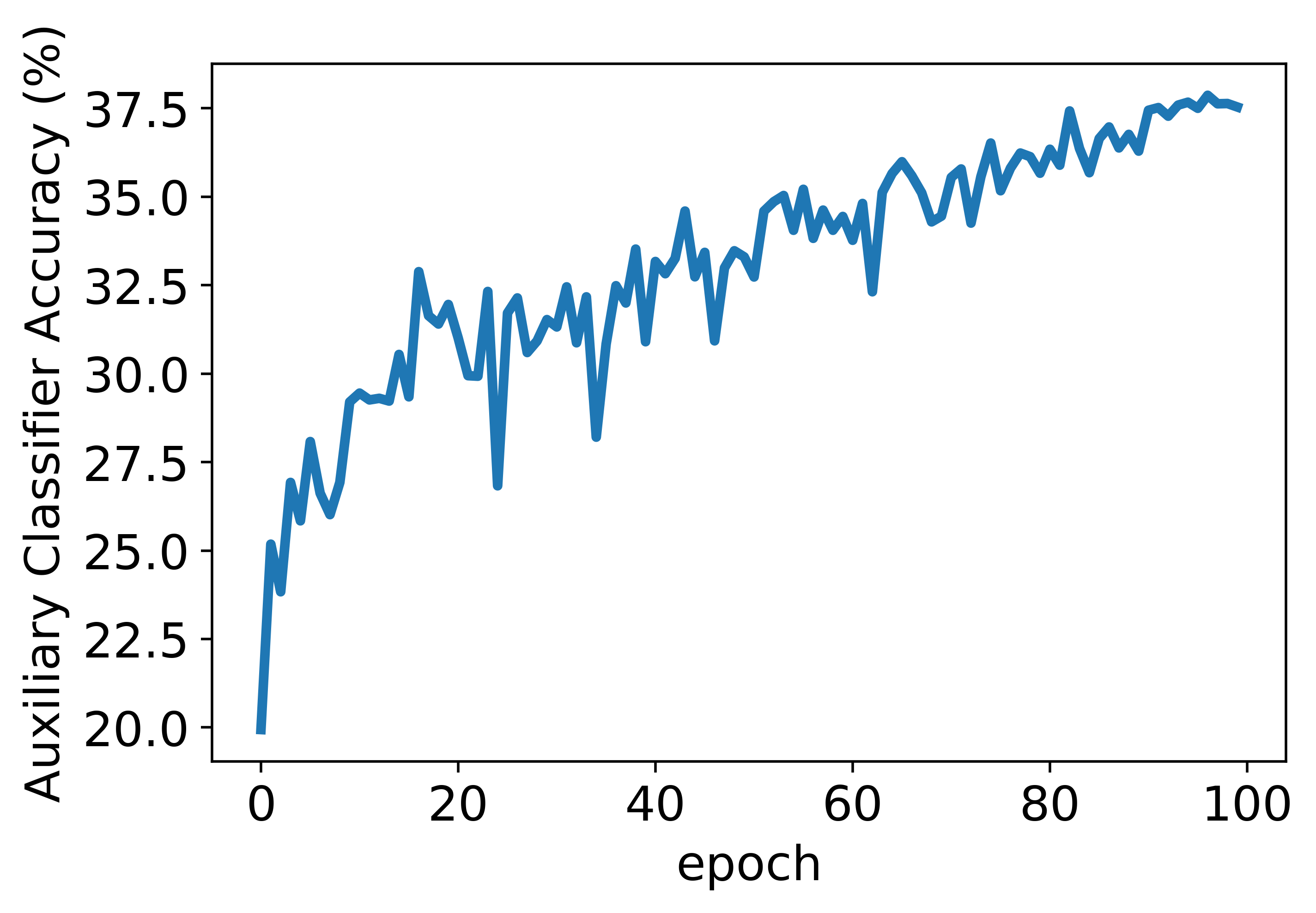}}
        \end{subfigure}
       \begin{subfigure}[CIFAR-100\label{fig:ssacc_cifar100}]{
        \includegraphics[width=0.47\linewidth]{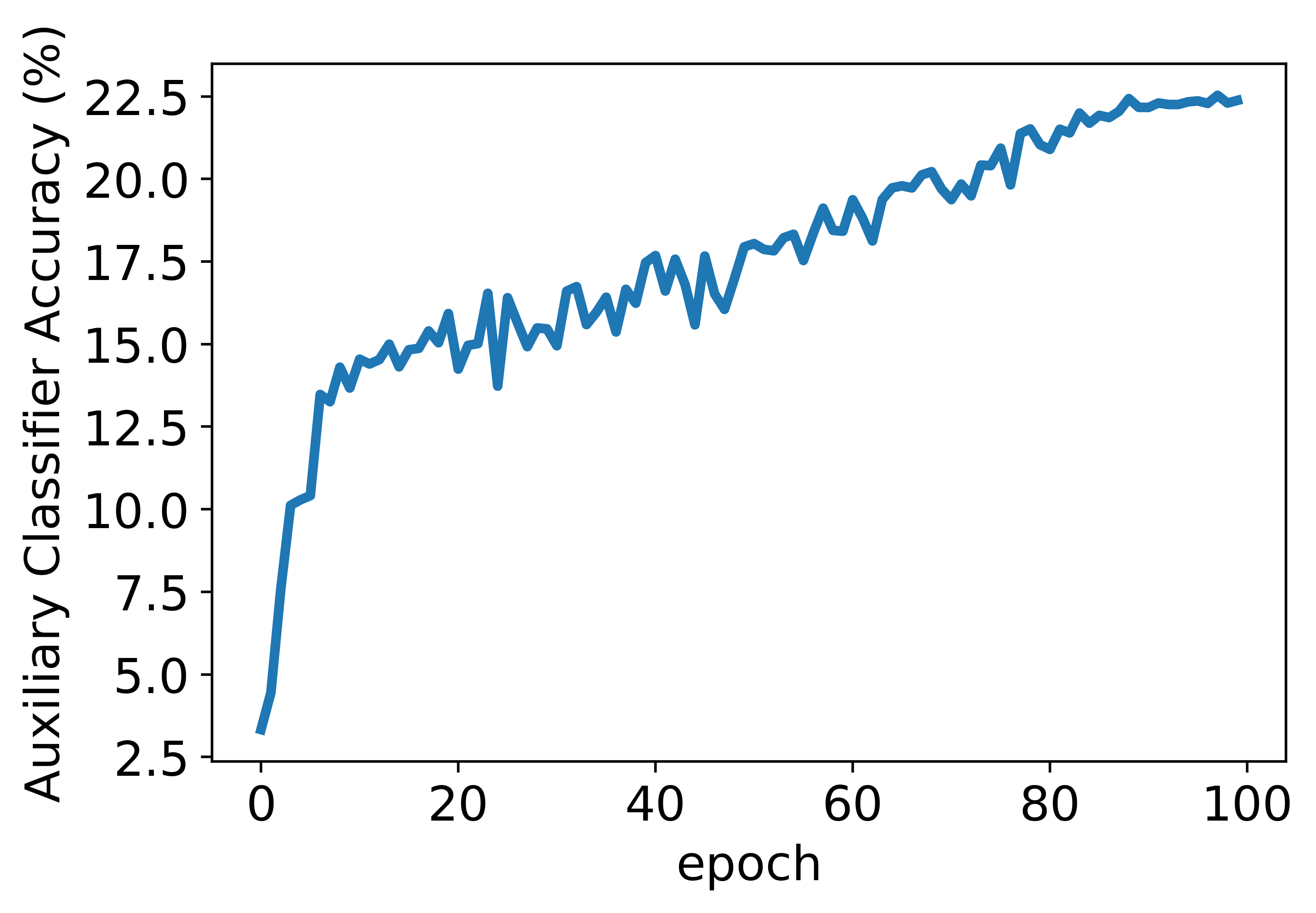}}
        \end{subfigure}
	\caption{Accuracy curves of the auxiliary classifier during knowledge transfer on WRN-40-2 \& WRN-16-1. }
	\label{fig:aux_acc}
\end{figure}


\subsection{Visualization}
We visualize synthetic images of our CSD from different training epochs in Figure~\ref{fig:vis}. We observe that images from early training epoch are more visually discernible than images from later training epoch, which indicates that as the number of training epochs increases, the student learning ability gradually becomes stronger, leading to more difficult synthetic images. Additionally, we plot the learning curves of the auxiliary classifier during knowledge transfer in Fig.~\ref{fig:aux_acc}. 

\begin{figure}[t]
    \centering
    \includegraphics[width=0.8\columnwidth]{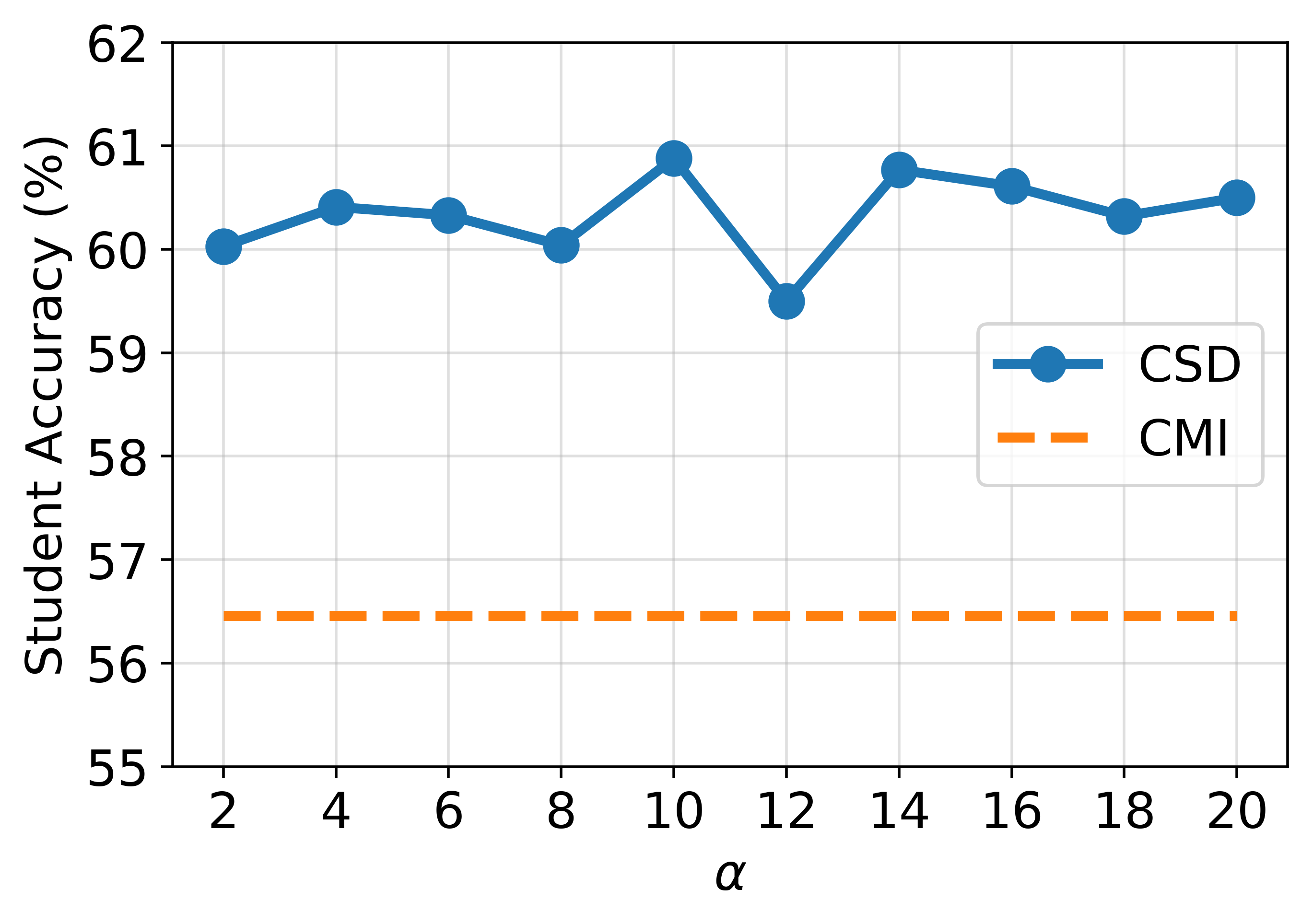}
    \caption{Sensitivity to hyper-parameter $\alpha$ on CIFAR-100 for WRN-40-2 \& WRN-16-1. The dash line refer to the mean student accuracy of CMI.}
    \label{fig:hyper}
\end{figure}

\subsection{Sensitivity Analysis}
To study how the hyper-parameter $\alpha$ affect the student final performance, we plot student accuracy curves on CIFAR-100 for WRN-40-2 \& WRN-16-1 with $\alpha$ ranging from 2 to 20 at equal interval of 2. From Fig.~\ref{fig:hyper}, we find that our CSD outperforms the best competitor (CMI) on all values of $\alpha$. 

\subsection{RELATED WORK}

\subsubsection{Data-Driven Knowledge Distillation}
Knowledge distillation (KD) is proposed to solve model compression problem by distilling knowledge from a cumbersome model (teacher) into a less-parameterized model (student). The vanilla KD \cite{hinton2015distilling} takes predictions from the last layer as the teacher knowledge to guide the student training. Besides predictions, many subsequent works excavate the knowledge in the output of intermediate layers to supervise the training of the student. The intermediate supervision can be formed by feature maps \cite{Romero2015FitNets,Chen2021Cross,chen2022knowledge,wang2023semckd}, attention maps \cite{Zagoruyko2017Paying, Tung2019SimilarityPreservingKD} or feature representation \cite{Tian2020ContrastiveRD}. There are also some works for transferring knowledge in relationships between different samples or layers \cite{Yim2017AGF,Tung2019SimilarityPreservingKD}. All the above mentioned methods are based on the premise that the original training data is available, while our proposed method is discussed in a more challenging scenario of no original data.

\subsection{Data-Free Knowledge Distillation}
Data-free knowledge distillation (DFKD) deals with transferring knowledge without the access to the original training data. A straightforward idea is to synthesize the original data for knowledge transfer. The approaches of data synthesis can be roughly categorized into two classes: inversion-based and generator-based approaches. Inversion-based approaches input the random Gaussian noise into the fixed teacher and update the input iteratively via the back-propogation until meeting certain constraints \cite{Yin2020DreamingTD,fang2021contrastive,fang2022up}. ADI \cite{Yin2020DreamingTD} proposes to leverage information stored in the batch normalization layers of the teacher to narrow gap between synthetic data and original data. CMI \cite{fang2021contrastive} introduces contrastive learning objective to address the mode collapse issue and thus ensure sample diversity. FastDFKD \cite{fang2022up} introduces a meta-synthesizer to  accelerate data synthesis process and achieves 100$\times$ faster speed. Generator-based approaches adopt a learnable generator to synthesize data \cite{Chen2019DataFreeLO,Binici2022RobustAR,Micaelli2019ZeroshotKT,choi2020data}. DAFL \cite{Chen2019DataFreeLO} introduce one-hot loss, activation loss and information entropy loss as the objective of synthesizing data, which are calculated according to the teacher output. PRE-DFKD \cite{Binici2022RobustAR} designs a Variational Autoencoder (VAE) to replay synthetic samples for preventing catastrophic forgetting without storing any data. Adversarial Distillation \cite{Micaelli2019ZeroshotKT,choi2020data} focus on synthesizing hard data by enlarging the divergence between predictions of the teacher and the student, so as to narrow the information gap between the teacher and the student. 

However, all above methods do not properly take into account the student's current ability during data synthesis, which may lead to oversimple samples and thus limit the final student performance.


\bibliographystyle{IEEEbib}
\bibliography{refs}

\bibliographystyle{IEEEbib}
\bibliography{refs}

\begin{thebibliography}{10}

\bibitem{simonyan2015Very}
Karen Simonyan and Andrew Zisserman,
\newblock ``Very deep convolutional networks for large-scale image
  recognition,''
\newblock in {\em ICLR}, 2015.

\bibitem{He2016Deep}
Kaiming He, Xiangyu Zhang, Shaoqing Ren, and Jian Sun,
\newblock ``Deep residual learning for image recognition,''
\newblock in {\em CVPR}, 2016, pp. 770--778.

\bibitem{Zagoruyko2016Wide}
Sergey Zagoruyko and Nikos Komodakis,
\newblock ``Wide residual networks,''
\newblock in {\em BMVC}, 2016.

\bibitem{hinton2015distilling}
Geoffrey~E Hinton, Oriol Vinyals, and Jeffrey Dean,
\newblock ``Distilling the knowledge in a neural network,''
\newblock {\em arXiv preprint arXiv:1503.02531}, 2015.

\bibitem{chen2022knowledge}
Defang Chen, Jian-Ping Mei, Hailin Zhang, Can Wang, Yan Feng, and Chun Chen,
\newblock ``Knowledge distillation with the reused teacher classifier,''
\newblock in {\em CVPR}, 2022, pp. 11933--11942.

\bibitem{Romero2015FitNets}
Adriana Romero, Nicolas Ballas, Samira~Ebrahimi Kahou, Antoine Chassang, Carlo
  Gatta, and Yoshua Bengio,
\newblock ``Fitnets: Hints for thin deep nets,''
\newblock in {\em ICLR}, 2015.

\bibitem{chen2020online}
Defang Chen, Jian-Ping Mei, Can Wang, Yan Feng, and Chun Chen,
\newblock ``Online knowledge distillation with diverse peers,''
\newblock in {\em AAAI}, 2020, pp. 3430--3437.

\bibitem{Chen2021Cross}
Defang Chen, Jian-Ping Mei, Yuan Zhang, Can Wang, Zhe Wang, Yan Feng, and Chun
  Chen,
\newblock ``Cross-layer distillation with semantic calibration,''
\newblock in {\em AAAI}, 2021, pp. 7028--7036.

\bibitem{wang2023semckd}
Can Wang, Defang Chen, Jian{-}Ping Mei, Yuan Zhang, Yan Feng, and Chun Chen,
\newblock ``Semckd: Semantic calibration for cross-layer knowledge
  distillation,''
\newblock {\em IEEE Transactions on Knowledge and Data Engineering}, vol. 35,
  no. 6, pp. 6305--6319, 2023.

\bibitem{Chen2019DataFreeLO}
Hanting Chen, Yunhe Wang, Chang Xu, Zhaohui Yang, Chuanjian Liu, Boxin Shi,
  Chunjing Xu, Chao Xu, and Qi~Tian,
\newblock ``Data-free learning of student networks,''
\newblock in {\em CVPR}, 2019, pp. 3514--3522.

\bibitem{Yin2020DreamingTD}
Hongxu Yin, Pavlo Molchanov, Jose~M Alvarez, Zhizhong Li, Arun Mallya, Derek
  Hoiem, Niraj~K Jha, and Jan Kautz,
\newblock ``Dreaming to distill: Data-free knowledge transfer via
  deepinversion,''
\newblock in {\em CVPR}, 2020, pp. 8715--8724.

\bibitem{fang2021contrastive}
Gongfan Fang, Jie Song, Xinchao Wang, Chengchao Shen, Xingen Wang, and Mingli
  Song,
\newblock ``Contrastive model inversion for data-free knowledge distillation,''
\newblock in {\em IJCAI}, 2021, pp. 2374--2380.

\bibitem{Yoo2019KnowledgeEW}
Jaemin Yoo, Minyong Cho, Taebum Kim, and U~Kang,
\newblock ``Knowledge extraction with no observable data,''
\newblock {\em NeurIPS}, 2019.

\bibitem{hao2021model}
Zhiwei Hao, Yong Luo, Zhi Wang, Han Hu, and Jianping An,
\newblock ``Model compression via collaborative data-free knowledge
  distillation for edge intelligence,''
\newblock in {\em ICME}, 2021, pp. 1--6.

\bibitem{Micaelli2019ZeroshotKT}
Paul Micaelli and Amos~J Storkey,
\newblock ``Zero-shot knowledge transfer via adversarial belief matching,''
\newblock {\em NeurIPS}, vol. 32, 2019.

\bibitem{choi2020data}
Yoojin Choi, Jihwan Choi, Mostafa El-Khamy, and Jungwon Lee,
\newblock ``Data-free network quantization with adversarial knowledge
  distillation,''
\newblock in {\em CVPR}, 2020, pp. 710--711.

\bibitem{fang2022up}
Gongfan Fang, Kanya Mo, Xinchao Wang, Jie Song, Shitao Bei, Haofei Zhang, and
  Mingli Song,
\newblock ``Up to 100x faster data-free knowledge distillation,''
\newblock in {\em AAAI}, 2022, pp. 6597--6604.

\bibitem{gidaris2018unsupervised}
Spyros Gidaris, Praveer Singh, and Nikos Komodakis,
\newblock ``Unsupervised representation learning by predicting image
  rotations,''
\newblock {\em ICLR}, 2018.

\bibitem{Yang2021HierarchicalSA}
Chuanguang Yang, Zhulin An, Linhang Cai, and Yongjun Xu,
\newblock ``Hierarchical self-supervised augmented knowledge distillation,''
\newblock in {\em IJCAI}, 2021.

\bibitem{Binici2022PreventingCF}
Kuluhan Binici, Nam~Trung Pham, Tulika Mitra, and Karianto Leman,
\newblock ``Preventing catastrophic forgetting and distribution mismatch in
  knowledge distillation via synthetic data,''
\newblock in {\em CVPR}, 2022, pp. 663--671.

\bibitem{Binici2022RobustAR}
Kuluhan Binici, Shivam Aggarwal, Nam~Trung Pham, Karianto Leman, and Tulika
  Mitra,
\newblock ``Robust and resource-efficient data-free knowledge distillation by
  generative pseudo replay,''
\newblock in {\em AAAI}, 2022.

\bibitem{Netzer2011ReadingDI}
Yuval Netzer, Tao Wang, Adam Coates, Alessandro Bissacco, Bo~Wu, and Andrew~Y
  Ng,
\newblock ``Reading digits in natural images with unsupervised feature
  learning,''
\newblock 2011.

\bibitem{krizhevsky2009learning}
Alex Krizhevsky, Geoffrey Hinton, et~al.,
\newblock ``Learning multiple layers of features from tiny images,''
\newblock 2009.

\bibitem{Sandler2018MobileNetV2}
Mark Sandler, Andrew Howard, Menglong Zhu, Andrey Zhmoginov, and Liang-Chieh
  Chen,
\newblock ``Mobilenetv2: Inverted residuals and linear bottlenecks,''
\newblock in {\em CVPR}, 2018, pp. 4510--4520.

\bibitem{chen2023geometric}
Defang Chen, Zhenyu Zhou, Jian-Ping Mei, Chunhua Shen, Chun Chen, and Can Wang,
\newblock ``A geometric perspective on diffusion models,''
\newblock {\em arXiv preprint arXiv:2305.19947}, 2023.

\bibitem{Zagoruyko2017Paying}
Nikos Komodakis and Sergey Zagoruyko,
\newblock ``Paying more attention to attention: improving the performance of
  convolutional neural networks via attention transfer,''
\newblock in {\em ICLR}, 2017.

\bibitem{Tung2019SimilarityPreservingKD}
Frederick Tung and Greg Mori,
\newblock ``Similarity-preserving knowledge distillation,''
\newblock in {\em CVPR}, 2019, pp. 1365--1374.

\bibitem{Tian2020ContrastiveRD}
Yonglong Tian, Dilip Krishnan, and Phillip Isola,
\newblock ``Contrastive representation distillation,''
\newblock in {\em ICLR}, 2020.

\bibitem{Yim2017AGF}
Junho Yim, Donggyu Joo, Jihoon Bae, and Junmo Kim,
\newblock ``A gift from knowledge distillation: Fast optimization, network
  minimization and transfer learning,''
\newblock in {\em CVPR}, 2017, pp. 4133--4141.

\end{thebibliography}

\end{document}